\documentclass[runningheads]{llncs}
\usepackage{graphicx}
\usepackage{xcolor}
\usepackage{subfigure}
\usepackage{placeins}
\usepackage{comment}
\usepackage{amsmath} 
\usepackage{amsfonts} 
\usepackage{hyperref} 
\usepackage[ruled,vlined,linesnumbered]{algorithm2e} 
\usepackage[misc]{ifsym} 

\begin{document}

\title{Task-agnostic Continual Hippocampus Segmentation for Smooth Population Shifts\thanks{Supported by the Bundesministerium für Gesundheit (BMG) with grant [ZMVI1-2520DAT03A].}}
\titlerunning{Task-agnostic Continual Hippocampus Segmentation}
%
\author{Camila Gonz\'{a}lez\inst{1}\orcidID{0000-0002-4510-7309}\textsuperscript{(\Letter)}
\and Amin Ranem\inst{1}
\and Ahmed Othman\inst{2}
\and Anirban Mukhopadhyay\inst{1}}

%
\authorrunning{Gonz\'{a}lez et al.}

%
\institute{Darmstadt University of Technology, Karolinenplatz 5, 64289 Darmstadt, Germany \email{camila.gonzalez@gris.tu-darmstadt.de} \and
University Medical Center Mainz, Langenbeckstraße 1, 55131 Mainz, Germany}

\maketitle              
\begin{abstract}

Most continual learning methods are validated in settings where task boundaries are clearly defined and task identity information is available during training and testing. We explore how such methods perform in a task-agnostic setting that more closely resembles dynamic clinical environments with gradual population shifts. We propose ODEx, a holistic solution that combines out-of-distribution detection with continual learning techniques. Validation on two scenarios of hippocampus segmentation shows that our proposed method reliably maintains performance on earlier tasks without losing plasticity.


\end{abstract}
\section{Introduction}

Deep learning methods are mostly validated in stationary environments where the train and test data have been carefully homogenized to preserve the i.i.d. assumption. This does not reflect the reality of clinical deployment, where acquisition conditions and disease patterns evolve over time. \emph{Continual learning} (CL) paradigms are being explored by medical imaging researchers \cite{ozgun2020importance,perkonigg2021continual,srivastava2021continual} and regulatory bodies \cite{vokinger2021regulating} as evaluation settings that are better suited for AI in healthcare. Continual methods deal with temporal restrictions on data availability by sequentially accumulating knowledge over a stream of \emph{tasks}, each containing data from a different distribution, without revisiting previous stages.

\begin{figure}
\includegraphics[width=\textwidth]{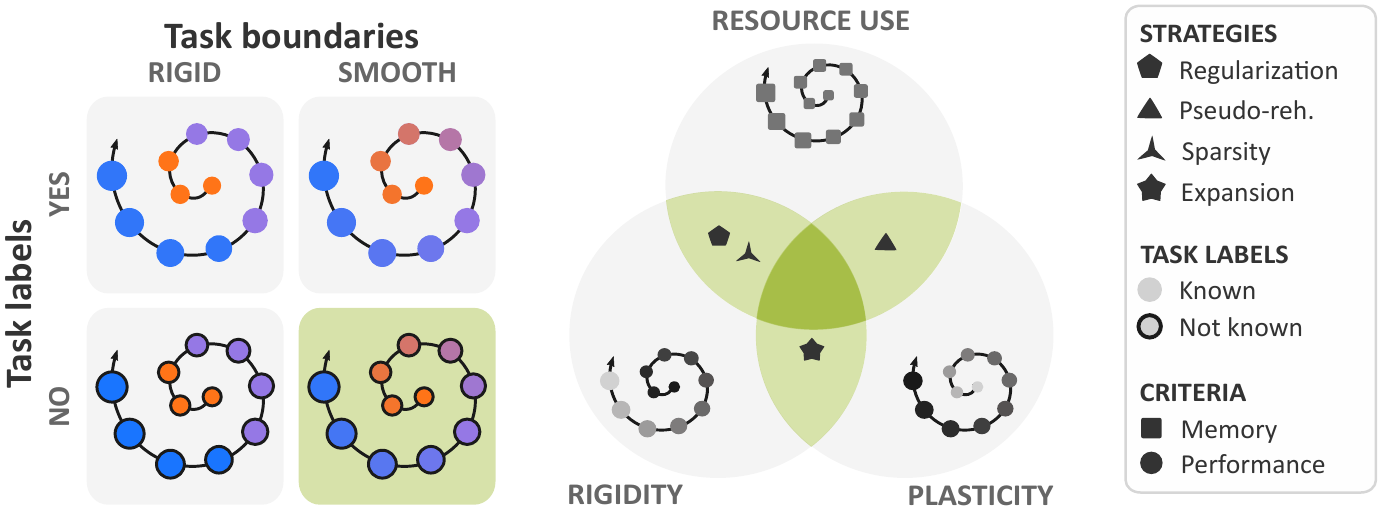}
\caption{Desiderata for continual learning \cite{delange2021continual,hadsell2020embracing}. Left: methods should not rely on rigid boundaries or task labels. Right: trade-off between plasticity, rigidity and resource use.}  \label{fig:requirements}
\end{figure}

Yet most CL approaches are validated in settings with \emph{rigid task boundaries} and \emph{known task labels}, which is far from how real dynamic environments behave \cite{delange2021continual}. When deviating from this simplistic problem formulation, they perform worse than simple baselines \cite{prabhu2020gdumb}. Previous research has established desirable properties for CL methods, illustrated in Fig. \ref{fig:requirements}. These include no reliance on either (1) assumptions on task boundaries during training or (2) access to task identity labels, i.e. the method should be \emph{task-agnostic} \cite{hadsell2020embracing}. In addition, the model should (3) preserve previous knowledge while (4) maintaining sufficient plasticity to learn new tasks and (5) not require additional computational resources during training \cite{delange2021continual,hadsell2020embracing}. The last three objectives are often deemed to be orthogonal, i.e. most approaches either \emph{catastrophically forget} previous knowledge (too plastic), cannot learn new tasks (too rigid) or the training time and resource requirements grow linearly with the number of tasks.

Methods for task-agnostic continual learning are overwhelmingly \emph{rehearsal-based} \cite{aljundi2019online,aljundi2019gradient,jin2021gradient,perkonigg2021dynamic,srivastava2021continual}, i.e. store a subset of past images or features in a memory buffer, which is not admissible in many diagnostic settings due to patient privacy considerations. \emph{Active learning} methods also exist which rely on expert interaction \cite{perkonigg2021continual}. 

Other approaches train generative models to identify distribution shifts \cite{rao2019continual} or only update the shortest sub-path of the network that allows a correct classification \cite{chen2020mitigating}, but such solutions are computationally expensive and are therefore only evaluated in low-resolution classification settings. The field of continual learning for medical segmentation is still under-studied. Most research follows regularization-based strategies that calculate the importance of parameters and penalize their deviation \cite{ozgun2020importance,zhang2021comprehensive}. Approaches have also been proposed for active learning \cite{zheng2021continual}, others allow the storage of previous samples \cite{perkonigg2021dynamic,venkataramani2019towards}. Some methods leverage feature disentanglement to alleviate forgetting \cite{lao2020continuous,memmel2021adversarial} 
 or maintain task-dependent batch normalization layers \cite{karani2018lifelong}. To our knowledge, no method has been previously introduced for semantic segmentation that is task-agnostic and does not make use of a rehearsal component.
  
 We propose \textbf{ODEx}, an expansion-based approach that (1) does not revisit previous stages, (2) is well-suited to a wide array of use cases, including semantic segmentation and (3) is task-agnostic, i.e. requires neither task boundaries nor task labels during training or inference. \emph{ODEx} uses continual out-of-distribution (OOD) detection to signal when to \emph{expand} the model and select the best parameters during inference. Although we maintain multiple parameter states in persistent memory, each occupies less than 0.2 GB and the continual OOD detection mechanism ensures that this number remains low. Unlike other methods, \emph{ODEx} requires the same GPU memory and training time as regular sequential learning. Our contributions include:
 \begin{enumerate}
     \item proposing a task-agnostic continual learning solution suitable for a wide array of deep learning architectures, and
     \item introducing a continual OOD detection mechanism that does not require access to early data for estimating the distance to the training distribution.
 \end{enumerate}

We explore the problem of hippocampus segmentation in T1-weighted MRIs, which is crucial for the diagnosis and treatment of neuropsychiatric disorders but highly sensitive to distribution shifts \cite{sanner2021reliable}, for two non-stationary environments. Our results show that \emph{ODEx} outperforms state-of-the-art approaches while adhering to desirable properties for continual learning.

\section{Methodology}

\begin{figure}
\includegraphics[width=1\textwidth]{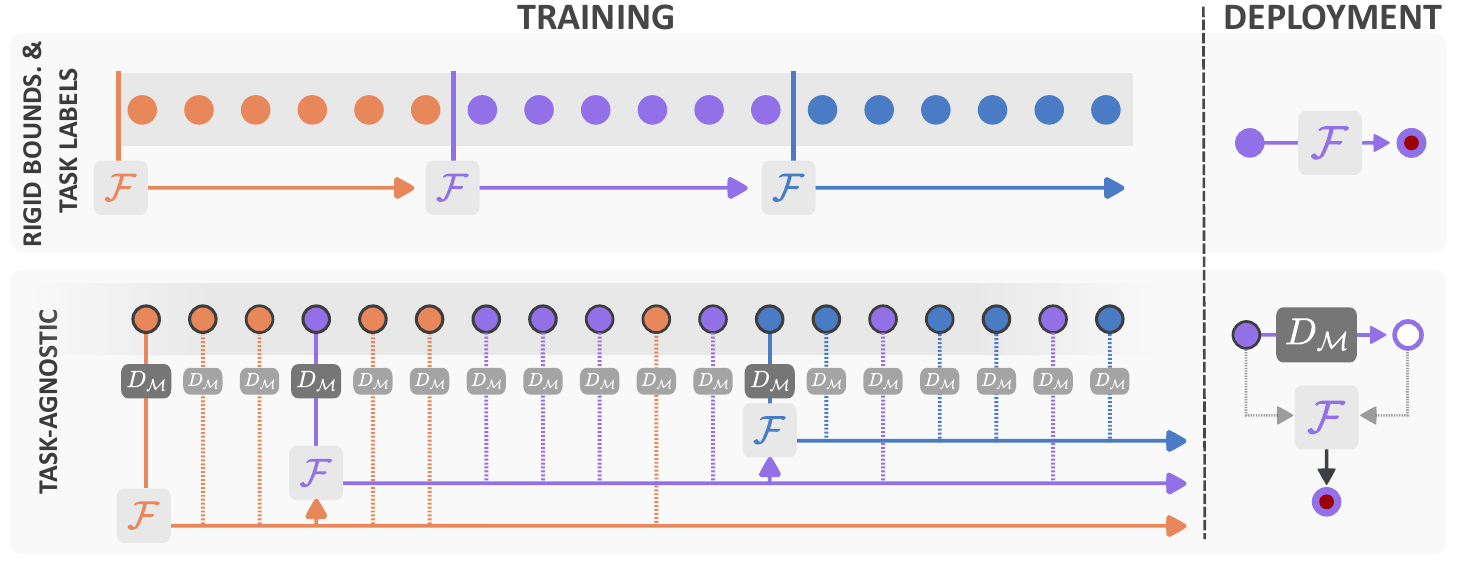}
\caption{Top: continual setting with rigid boundaries and task labels. Expansion methods create new parameters at each task boundary. 
Bottom: the task-agnostic \emph{ODEx} method initializes a new set of parameters when a domain shift is detected.} \label{fig:lifelong_learning}
\end{figure}

We start by defining our problem formulation of task-agnostic continual learning. We then introduce \emph{ODEx},
visualized in Fig. \ref{fig:lifelong_learning} (bottom). During training, we accumulate the mean and covariance of batch normalization layers and detect domain shifts with the Mahalanobis distance.
When a domain shift occurs, a new model is initialized with the most appropriate parameters and added to the model pool.
During inference, we extract predictions with the best model state.

\textbf{Task-agnostic continual learning:} In continual learning settings, model $\mathcal{F}_\theta : x \rightarrow \hat{y}$ is trained with data samples from an array of ${N_t}$ different \emph{tasks} or data distributions $\left \{ \mathcal{T}_i ... \mathcal{T}_{N_t} \right \}$, each found at the $i_{th}$ \emph{stage} $t_i$. The model should be deployable after finishing the first stage, and evolve over time. For segmentation, each instance has the form $(x, y, j)$, where $x$ is an image and $y$ the segmentation mask. Additionally, $j$ denotes the \emph{task label}, i.e. that $(x, y)\ \sim\ \mathcal{T}_j$. The goal is to find parameters $\theta$ that minimize the loss $\mathcal{L}$ over all seen tasks $\left \{ \mathcal{T}_i \right \}_{i \leq N_{t}}$ (Eq. \ref{eq:min_loss}).

\begin{equation} \label{eq:min_loss}
\underset{\theta}{\arg\min}\sum_{j=1}^{N_{t}}\mathbb{E}_{(x, y)\ \sim\ \mathcal{T}_j}\ \left [ \mathcal{L}\left ( \mathcal{F}_\theta(x),y \right )\right ]
\end{equation}

The objective cannot be optimized directly, as at any training stage $t_j$ only data from $\mathcal{T}_{j}$ is available. The main challenge consists of ensuring enough \emph{rigidity} during training to obtain good performance on $(x, y) \sim \left \{ \mathcal{T}_{i} \right \}_{i<j}$ and enough \emph{plasticity} to learn from present and future data $(x, y) \sim \left \{ \mathcal{T}_{i} \right \}_{i \geq j}$. 

\emph{Expansion-based methods} approach this by keeping \emph{task-dependent} parameters $\left \{ \theta_1 ... \theta_{N_t} \right \}$, which in their simplest form comprise the entire model, and perform inference on $(x, y, j)$ with the respective $\mathcal{F}_{\theta_j}$ (see Fig. \ref{fig:lifelong_learning}, above). 
In \emph{task-agnostic scenarios}, task labels $j$ are unknown and may not even be clearly defined. The goal is to learn a set of parameters $\Theta = \left \{ \theta_1 ... \theta_{|\Theta|} \right \}$ and an inference function $\mathcal{J}: x \rightarrow \theta$ that selects the best parameters during testing (Eq. \ref{eq:min_loss_agnostic}). In the absence of rigid task boundaries, the size of the model pool $|\Theta|$ is unknown. Task-agnostic settings thus signify three additional challenges: (1) detecting when domain shifts occur, (2) keeping $|\Theta|$ low and (3) choosing the best parameters during testing. In the following, we outline how we approach these.

\begin{equation} \label{eq:min_loss_agnostic}
\underset{\Theta}{\arg\min}\sum_{j=1}^{N_{t}}\mathbb{E}_{(x, y)\ \sim\ \mathcal{T}_j}\ \left [ \mathcal{L}\left ( \mathcal{F}_{\mathcal{J}(x)}(x),y \right )\right ]
\end{equation}

\textbf{Detecting domain shifts:} During training, we extract features $z$ from the first set of \textit{Batch Normalization} layers $BN_1$. These normalize inputs and thus contain domain-pertinent information which has been found to play a key role in detecting interference during sequential learning \cite{karani2018lifelong}. We estimate a multi-variate Gaussian $\mathcal{N}_i(\mu_i, \Sigma_i)$  at the end of training stage $t_i$ as:

\begin{equation} \label{eq:batch_norm}
    z_k \leftarrow BN_1(x_k); \phantom{--} \mu_i \leftarrow \frac{1}{N} \sum_{k=1}^N z_k; \phantom{--} \Sigma_i \leftarrow \frac{1}{N} \sum_{k=1}^N (z_k - \mu_i)(z_k - \mu_i)^T
\end{equation}

Inspired by previous research on OOD detection for semantic segmentation \cite{gonzalez2021detecting}, we detect data shifts by calculating the \emph{Mahalanobis distance} $D_\mathcal{M}(z;\mu, \Sigma)$ to the training distribution. In contrast to other methods for assessing similarity, such as the \emph{Gram distance} popular in rehearsal-based continual learning \cite{perkonigg2021dynamic,perkonigg2021continual}, the Mahalanobis distance requires storing only $\mu$ and $\Sigma$.

As we cannot revisit data from previous stages, we cannot estimate $\mathcal{N}$ with all data used to train the model. In a situation with slowly shifting data distributions, if we were to only consider the $\mu$ and $\Sigma$ of the last training batch, then we may never detect a sufficiently large distance signaling the need to expand the model pool. We therefore store $\mu_i$ and $\Sigma_i$ at the end of each training stage $t_i$ and add this to the \emph{history} $\mathcal{B}_i$ of the model which contains information from all pertinent training stages. At stage $t_{i+1}$, parameters $\hat{\theta}$ are selected that minimize the summed distance of the present training data to the history of $\hat{\theta}$ (Eq. \ref{eq:best_theta}).

\begin{equation} \label{eq:best_theta}
    D_\mathcal{M}(z; i): \underset{\theta_j \in \Theta_{i}}{\min} \sum_{(\mu_j, \Sigma_j) \in \mathcal{B}_j} D_{\mathcal{M}_j}(z;\mu_j, \Sigma_j)
\end{equation}

\textbf{Managing the model pool: } When data arrives for a new stage $t_{i}$, the distance $D_\mathcal{M}(z; i)$ is calculated and the best model $\hat{\theta}$ is selected. If $D_\mathcal{M}(z; i) < \xi$ (case 1), then $\hat{\theta}$ is updated with the current data. Afterwards, $\mu_i$ and $\Sigma_i$ are calculated and added to the model history $\hat{\mathcal{B}}$. If instead $D_\mathcal{M}(z; i) \geq \xi$ (case 2), a domain shift is detected and a new model $\theta_i$ is initialized with the parameters of $\hat{\theta}$. After a domain shift, the size of the model pool $|\Theta|$ grows by 1. The history of the new model $\mathcal{B}_i$ is initialized with $\hat{\mathcal{B}}$, so the history of each model contains information pertaining to all data distributions used to train it. Following previous research \cite{gonzalez2021detecting} we normalize the distances between the minimum and doubled maximum in-distribution values, and set $\xi = 2\mu$.

Continuing to train older models instead of initializing a new one for each stage has two advantages: (1) the model pool does not grow linearly with the length of the data stream, which would be prohibiting for deployment over long time periods and (2) models can benefit from further training when the data distributions are compatible, potentially allowing positive backwards transfer.

\textbf{Performing inference: } Inference proceeds as illustrated in Fig. \ref{fig:lifelong_learning} (right). For each image, the summed Mahalanobis distance of the test image to each set of parameters $\theta \in \Theta$ is calculated. Again, the best model $\hat{\theta}$ is selected and, in this case, directly used to extract a segmentation mask $\mathcal{F}_{\hat{\theta}}(x) = \hat{y}$. 

\section{Experimental Setup}

We briefly outline how we build our data base of tasks with smooth distribution shifts from publicly available datasets and report relevant aspects of our experimental setup. For further implementation details, we refer the reader to the supplementary material and our code found under \url{https://github.com/MECLabTUDA/Lifelong-nnUNet}.

\textbf{Data:} We look at two different scenarios of data streams with slowly shifting distributions for segmentation of the entire hippocampus (head, body and tail) in T1-weighted MRIs. The first is constructed from three public datasets: \emph{HarP} \cite{boccardi2015training} contains 135 healthy and Alzheimer’s disease patients, \emph{Dryad} \cite{kulaga2015multi} has 25 healthy adult subjects and \emph{Decathlon} \cite{medicaldecathlon} contains 130 healthy and schizophrenia patients. We slowly shift the distribution of cases from each source as illustrated in Appendix A. We refer to this scenario as \textbf{shifting source}. For the second scenario, henceforth referred to as \textbf{transformed}, we slowly modify the \emph{Decathlon} data using the \emph{TorchIO} library \cite{perez-garcia_torchio_2021}. We apply intensity rescaling up to a contrast stretching of (0.1, 0.9) and affine transformations of up to a (0.8, 1.2) scaling range, 15 degrees rotation and 5 mm translation.

\textbf{Network architecture and training:} We use a full-resolution \emph{nnUNet} \cite{isensee2021nnu} model for all experiments, with the architecture and training settings selected for the first training stage of each data stream. We perform 200 epochs for each stage, with a loss of \emph{Dice} and \emph{Binary Cross Entropy} weighted equally. All experiments were carried out on a \emph{Nvidia Tesla T4} GPU (16 GB).

\textbf{Metrics:} We report the average Dice on test data from all tasks $\left \{ \mathcal{T}_i \right \}_{i \leq N_{t}}$ as well as backwards (BWT) and forwards (FWT) transferability \cite{delange2021continual,hadsell2020embracing}. BWT is the \emph{inverse forgetting} and displays to what extent the performance on test samples $(x, y) \sim \mathcal{T}_i$  deteriorates with further training in stages $\left \{ t_i \right \}_{i > N_{t}}$. FWT instead measures what impact training on each stage $\left \{ t_i \right \}_{i \leq N_{t}}$ has on test data $(x, y) \sim \mathcal{T}_i$. Methods that prevent forgetting show high, realistically close to 0, BWT. FWT is high if enough plasticity is maintained to acquire new knowledge. For both metrics, we report the average over test data from all tasks.

\textbf{Baselines:} In Sec. \ref{sec:cl_performance}, we compare our approach against sequential training and five popular continual learning approaches: Elastic Weight Consolidation (EWC) \cite{kirkpatrick2017overcoming}, Modelling the Background (MiB) \cite{cermelli2020modeling}, Riemannian Walk (RW) \cite{chaudhry2018riemannian}, PLOP \cite{douillard2021plop} and Learning without Forgetting (LwF) \cite{li2017learning}. We also report the upper bound of joint training. In most cases, we use the hyperparameters suggested in the corresponding publications or code bases (for more details see Appendix B). For MiB, we reduce the \emph{lkd} to prevent loss explosion. In Sec. \ref{sec:ablation} we perform an ablation study and compare the use of the Mahalanobis distance to other methods proposed within task-agnostic learning, namely using the Gram matrix \cite{perkonigg2021dynamic} and detecting domain shifts through a fall in training performance \cite{chen2020mitigating}.

\section{Results}

\setlength{\tabcolsep}{2pt}
\begin{table}[t]
\centering
\caption{Performance of the joint training upper bound (first row), sequential learning and six continual learning strategies on the two hippocampus segmentation scenarios. }\label{tab:method_against_sota}
\begin{tabular}{l|lll|lll}
& \multicolumn{3}{c}{\textbf{Shifting source}} & \multicolumn{3}{c}{\textbf{Transformed}}\\
Method & Dice $\uparrow$ & BWT $\uparrow$ & FWT $\uparrow$ & Dice $\uparrow$ & BWT $\uparrow$ & FWT $\uparrow$\\
\hline
\hline
Joint & .89 $\pm$.01 & & & .90 $\pm$.01 & & \\
\hline
   Seq.  & .57 $\pm$.32  & -.19 $\pm$.12  & \textbf{.14} $\pm$.09  & .87 $\pm$.03 & -.02 $\pm$.02  & .09 $\pm$.05\\
    EWC  & .78 $\pm$.08   & \textbf{\phantom{-}.02} $\pm$.03  & .08 $\pm$.08  & .79 $\pm$.10 & \textbf{\phantom{-}.01} $\pm$.01  & .04 $\pm$.02  \\
    MiB  & .67 $\pm$.24  & -.10 $\pm$.07  & \textbf{.14} $\pm$.10  & .87 $\pm$.04 & -.02 $\pm$.02  & .07 $\pm$.04 \\
     RW  & .61 $\pm$.28  & -.15 $\pm$.10  & \textbf{.14} $\pm$.10  & .87 $\pm$.03 & -.03 $\pm$.03  & .09 $\pm$.05 \\
   PLOP  & .57 $\pm$.32  & -.22 $\pm$.14  & .13 $\pm$.09  & .86 $\pm$.02 & -.02 $\pm$.02  & \textbf{.10} $\pm$.06 \\  
    LwF  & .51 $\pm$.35  & -.23 $\pm$.13  & .10 $\pm$.07  & .86 $\pm$.04 & -.04 $\pm$.04  & \textbf{.10} $\pm$.06 \\   
   ODEx (ours) & \textbf{.87} $\pm$.04  & -.03 $\pm$.02  & \textbf{.14} $\pm$.09  & \textbf{.89} $\pm$.01 & -.01 $\pm$.01  & .09 $\pm$.05 \\ 
\end{tabular}
\end{table}

We first compare \emph{ODEx} to state-of-the-art continual learning approaches in Sec. \ref{sec:cl_performance}. Afterwards, we take a closer look at the cumulative Mahalanobis distance for identifying domain shifts and selecting the best parameters (Sec. \ref{sec:ablation}).

\subsection{Continual learning performance} \label{sec:cl_performance}

\begin{figure}
\includegraphics[width=\textwidth]{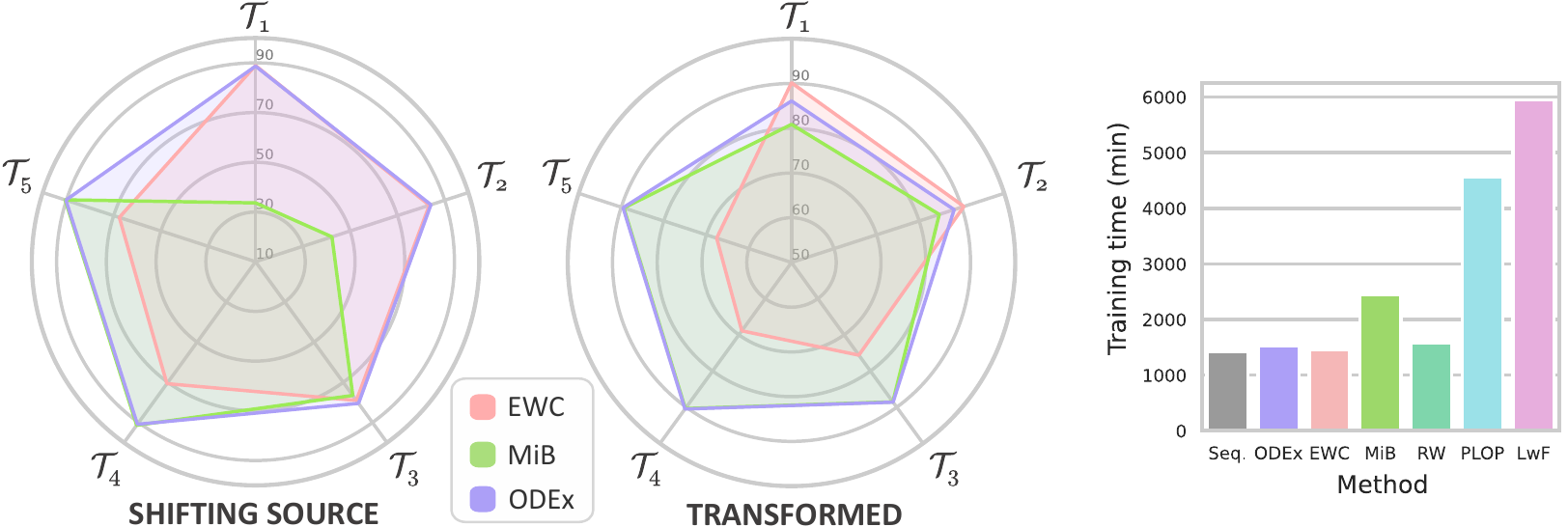}
\caption{Left: Per-task Dice. EWC and MiB are at opposite ends of the plasticity/rigidity spectrum, whereas ODEx allows for further training without compromising performance on previous tasks. Right: training times for the shifting source scenario.} \label{fig:spiders}
\end{figure}

We compare our proposed approach \emph{ODEx} to five continual learning methods in Tab. \ref{tab:method_against_sota}. The first row shows the upper bound of training a model statically with all training data. \emph{Sequential} results show the deterioration of the performance in earlier tasks as training is carried out, and the following rows display how five continual learning strategies alleviate this. From these, only \emph{EWC} maintains performance on earlier tasks, but at the cost of losing model plasticity and being unable to acquire new knowledge. \emph{ODEx} instead reaches a high FWT showing effective learning on later tasks while still performing well on data from the first training stages. This behavior is further illustrated in Fig. \ref{fig:spiders} (left), where the per-task performance is plotted for \emph{EWC}, which successfully retains old knowledge, \emph{MiB}, which reaches a high Dice on later tasks, and \emph{ODEx} that performs well on data from all stages. This is particularly clear for the more difficult \emph{shifting source} case, but a Wilcoxon one-sided signed-rank test affirms that \emph{ODEx} significantly outperforms all other approaches in terms of Dice score for both scenarios.

As for resource utilization, \emph{ODEx} requires no more GPU memory than sequential training, as we update one model at a time. The estimation of $\Sigma$ and the calculation of $D_\mathcal{M}$ can be carried out in the CPU given the low resolution of $z$. Fig. \ref{fig:spiders} (right) shows that \emph{ODEx} takes only marginally longer than training without any method for forgetting prevention. Though several models are stored (two for \emph{shifting source} and four for \emph{transformed}, see Tab. \ref{tab:ablation_study}) each weights less than 200 MB, being far from a limiting factor in practice.

\begin{figure}[h]
\centering
\includegraphics[width=0.92\textwidth]{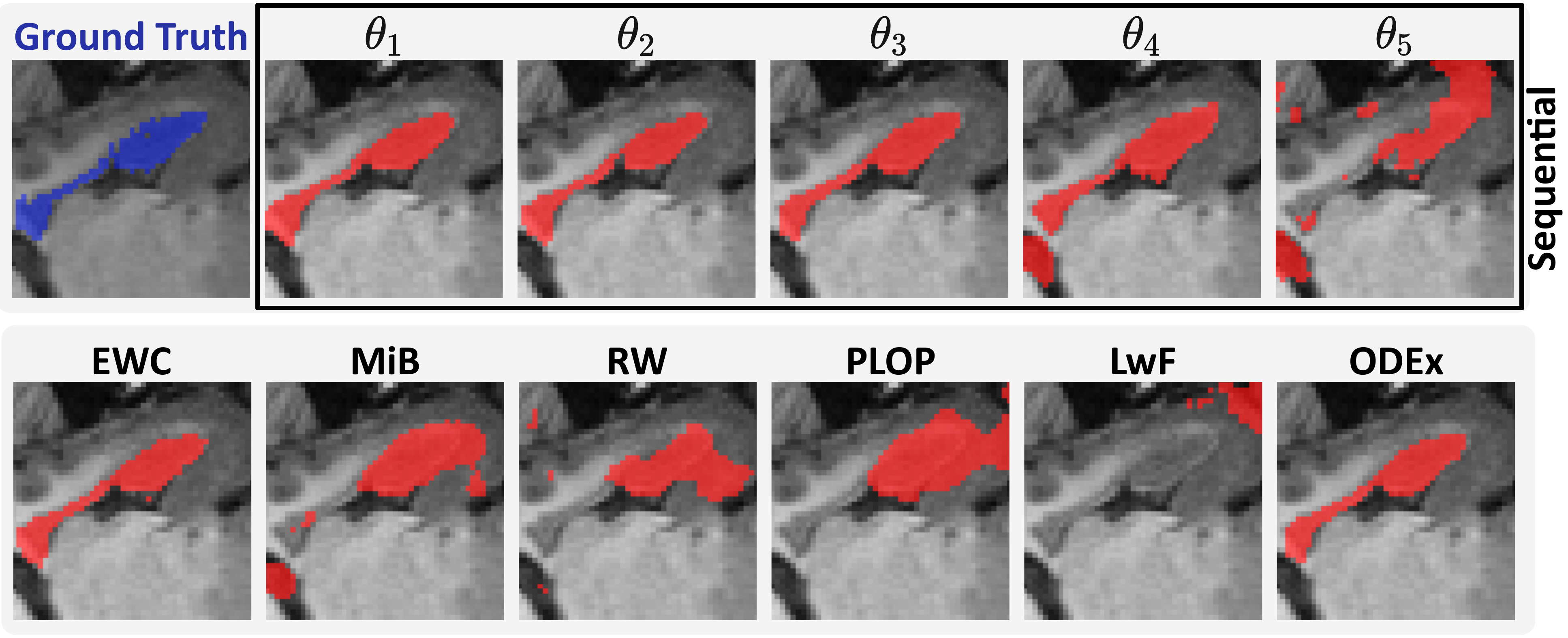}
\caption{Crops with overlayed segmentations for axial slice 25 of a subject from $\mathcal{T}_1$ (shifting source). Top: ground truth (blue) and performance deterioration with regular SGD. Bottom: six continual learning methods, after finishing training on last stage.} \label{fig:qualitative}
\end{figure}

Fig. \ref{fig:qualitative} qualitatively shows in the upper row the sequential deterioration of the segmentation for a test subject $(x, y) \sim \mathcal{T}_1$. The lower row displays the segmentation masks produced by each continual learning method. Though the head is mostly segmented well by several methods, only \emph{EWC} and \emph{ODEx} properly segment the body and tail and maintain the integrity of the shape.

\subsection{Ablation study} \label{sec:ablation}

In Tab. \ref{tab:ablation_study} we compare our strategy for detecting when to grow the model pool to previous work in the field of task-agnostic learning. The performance of all methods is very similar for the easier \emph{transformed} scenario, but we see clear differences in \emph{shifting source}. We first explore two versions of \emph{ODEx} that use our proposed strategy for selecting the best model but detect domain shifts in a different fashion. \emph{ODEx} $-\infty\ \xi$ creates a new model for every stage. The lower Dice suggests that the models suffer from the lack of training data, and $|\Theta|$ grows linearly with the number of training stages. \emph{DiceEx} initializes a new model when the training Dice falls more than 10\%, which results in higher forgetting. \emph{ODEx} ${\displaystyle \neg} \mathcal{B}$ shows the situation where we do not keep a history for the training distributions of previous stages and only calculate the distance to the last stage. For this version, no new model is initialized for \emph{shifting source} and the single available model significantly forgets previous knowledge. Finally, we test the use of the Gram distance instead of Mahalanobis for both training and testing, and find that it does not properly detect distribution shifts for \emph{shifting source}.

\setlength{\tabcolsep}{2pt}
\begin{table}
\centering
\caption{Performance of different strategies for detecting domain boundaries and/or selecting a model state during inference.}\label{tab:ablation_study}
\begin{tabular}{l|lllc|lllc}
& \multicolumn{4}{c}{\textbf{Shifting source}} & \multicolumn{4}{c}{\textbf{Transformed}}\\
Method & Dice $\uparrow$ & BWT $\uparrow$ & FWT $\uparrow$ & $\left | \Theta \right |$$\downarrow$ & Dice $\uparrow$& 
BWT $\uparrow$ & FWT $\uparrow$ &  $\left | \Theta \right |$$\downarrow$\\
\hline
\hline
   ODEx (ours) & \textbf{.87} $\pm$.04  & -.03 $\pm$.02  & \textbf{.14} $\pm$.09& 2 & .89 $\pm$.01  & -.01 $\pm$.01  & \textbf{.09} $\pm$.05& 4 \\ 
   
   ODEx $-\infty\ \xi$  & .83 $\pm$.04  & \textbf{\phantom{-}.00} $\pm$.00  & .11 $\pm$.09 & 5 & .89 $\pm$.01  & \textbf{\phantom{-}.00} $\pm$.00  & \textbf{.09} $\pm$.05 & 5\\
   
   DiceEx \cite{chen2020mitigating}  & .84 $\pm$.08  & -.07 $\pm$.03  & \textbf{.14} $\pm$.10 & 2 & .89 $\pm$.02  & -.01 $\pm$.01  & \textbf{.09} $\pm$.05 & \textbf{2}\\   
   
   ODEx ${\displaystyle \neg} \mathcal{B}$ \cite{gonzalez2021detecting}  & .57 $\pm$.32  & -.19 $\pm$.12  & \textbf{.14} $\pm$.09 & \textbf{1} & .89 $\pm$.01   & \textbf{\phantom{-}.00} $\pm$.00  & \textbf{.09} $\pm$.05  & 3\\

Gram  \cite{perkonigg2021dynamic}& .57 $\pm$.32  & -.19 $\pm$.12  & \textbf{.14} $\pm$.09 & \textbf{1} & \textbf{.90} $\pm$.01   & \textbf{\phantom{-}.00} $\pm$.00  & \textbf{.09} $\pm$.05 & 3
\end{tabular}
\end{table}

\section{Conclusion}

We introduce \emph{ODEx}, an expansion-based continual learning strategy suitable for real clinical environments with smooth acquisition and population shifts. We evaluate our approach on two hippocampus segmentation scenarios and show that it outperforms state-of-the-art methods by maintaining good performance on data from early stages without compromising model plasticity. \emph{ODEx} requires only marginally higher training times than regular sequential learning, and the same amount of GPU memory. While additional persistent storage is needed to store different sets of parameters, the OOD detection strategy keeps this number low. Each explored scenario required less than 0.8 GB, rendering this limitation insignificant in practice. Future work should explore whether it suffices to maintain only a subset of domain-specific parameters, such as the last decoder blocks or batch normalization layers. By releasing our code and models, we hope to boost continual learning research in task-agnostic medical settings.

%
%
\bibliographystyle{splncs04}
\bibliography{bibliography}

\end{document}


\title{Task-agnostic Continual Hippocampus Segmentation for Smooth Population Shifts\thanks{Acknowledgements withheld.}}
\titlerunning{Task-agnostic Continual Hippocampus Segmentation}
%

%
\authorrunning{Gonz\'{a}lez et al.}

%
\institute{Anonymous institution(s)}

%

\appendix 

\section{Data scenarios} \label{sec:annex_data}

\begin{figure}
\includegraphics[width=\textwidth]{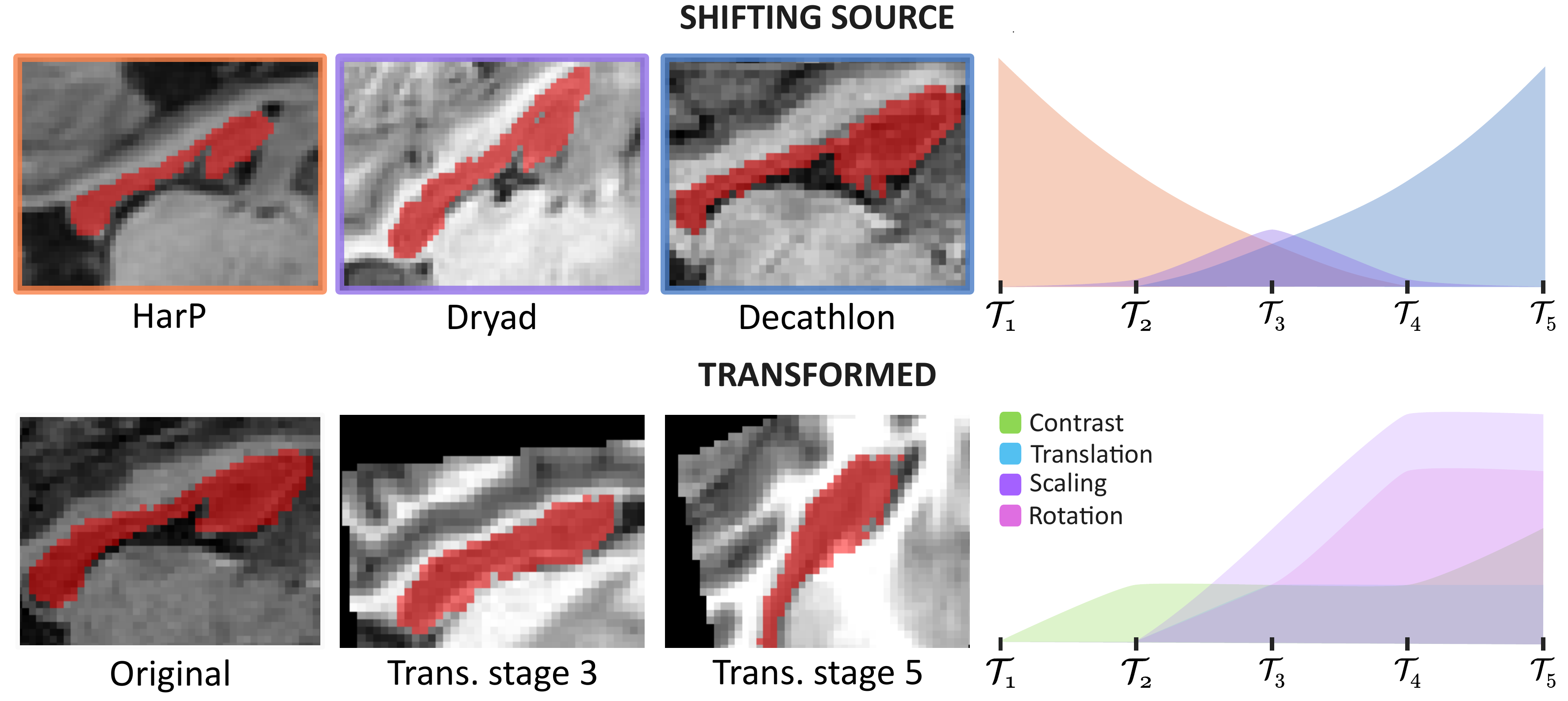}
\caption{The two scenarios of data streams with distribution shifts explored in this work. Top: number of cases from three datasets is slowly shifted. Bottom: the \emph{Decathlon} dataset is artificially transformed. We used the first 80/20 split generated by the nnUNet framework for \emph{HarP}, \emph{Dryad} and \emph{Decathlon} and ensured that test cases remained as such across both scenarios.} \label{fig:data}
\end{figure}

\section{Architecture and training parameters}

\begin{table}
\centering
\caption{Hyperparameters for training continual learning methods. The settings specified in the first row were used for all experiments.}
\begin{tabular}{p{2cm}|p{10cm}}

\textbf{Method} & \textbf{Setting} \\
\hline
\hline
All & $\text{optimizer}=\text{SGD}$, $\text{lr}=0.01$, $\text{weight decay}=3e-5$, $\text{momentum}=.99$, $\text{nr. blocks}=4$ for \emph{shifting source}, $\text{nr. blocks}=3$ for \emph{transformed} \\
\hline
EWC & $\lambda=0.4$ \\
\hline
MiB & $\alpha=0.9$, $\text{lkd}=1$ for \emph{shifting source}, $\text{lkd}=0.1$ for \emph{transformed} \\
\hline
RW & $\alpha=0.9$, $\lambda=0.4$, $\text{update after}=10$\\
\hline
PLOP & $\lambda=0.01$, $\text{scales}=3$, resampling to (48, 48, 48) for \emph{transformed}, no resampling for \emph{shifting source}\\
\hline
LwF & $T=2$ \\
\hline
\end{tabular}
\end{table}

\newpage
\section{Calculation of evaluation metrics}

Considering $\mathcal{F}_i(x)=\hat{y}_i$ as the prediction made at stage $t_i$, backwards transfer (BWT) is the change in performance after training with each subsequent task $\left \{ \mathcal{T} \right \}_{j>i}$, averaged over the number of samples in $\mathcal{T}_i$ and the number of tasks (Eq. \ref{eq:bwt}). BWT is not defined for the last task $\mathcal{T}_{N_t}$, as $\left \{ t_j \right \}_{j>N_t}= \emptyset$.

\begin{equation} \label{eq:bwt}
    BWT = \frac{1}{N_t}\sum_{i=1}^{N_t} \left [    \frac{1}{ \left | \left \{ t_j \right \}_{j>i} \right | } \sum_{j>i}  \left [ \frac{1}{\left | \mathcal{T}_i \right |} \sum_{k=1}^{\left | \mathcal{T}_i \right |}\text{Dice}\phantom{.}(\mathcal{F}_j(x_k), y_k) - \text{Dice}\phantom{.}(\mathcal{F}_i(x_k), y_k))
 \right ] \right ]
\end{equation}

Forwards transfer (FWT) is, for each task $\mathcal{T}_i$, the change in performance in each stage before and up to $t_i$, averaged over the number of samples and tasks. FWT is not defined for the first task $\mathcal{T}_1$, as $\left \{ t_j \right \}_{j<1}= \emptyset$.

\begin{equation}
    FWT = \frac{1}{N_t}\sum_{i=1}^{N_t} \left [    \frac{1}{ \left | \left \{ t_j \right \}_{j\leq i} \right | } \sum_{j \leq i}  \left [ \frac{1}{\left | \mathcal{T}_i \right |} \sum_{k=1}^{\left | \mathcal{T}_i \right |}\text{Dice}\phantom{.}(\mathcal{F}_{j}(x_k), y_k) - \text{Dice}\phantom{.}(\mathcal{F}_{j-1}(x_k), y_k))
     \right ] \right ]
\end{equation}

\section{Static learning results}

\begin{figure}
\includegraphics[width=\textwidth]{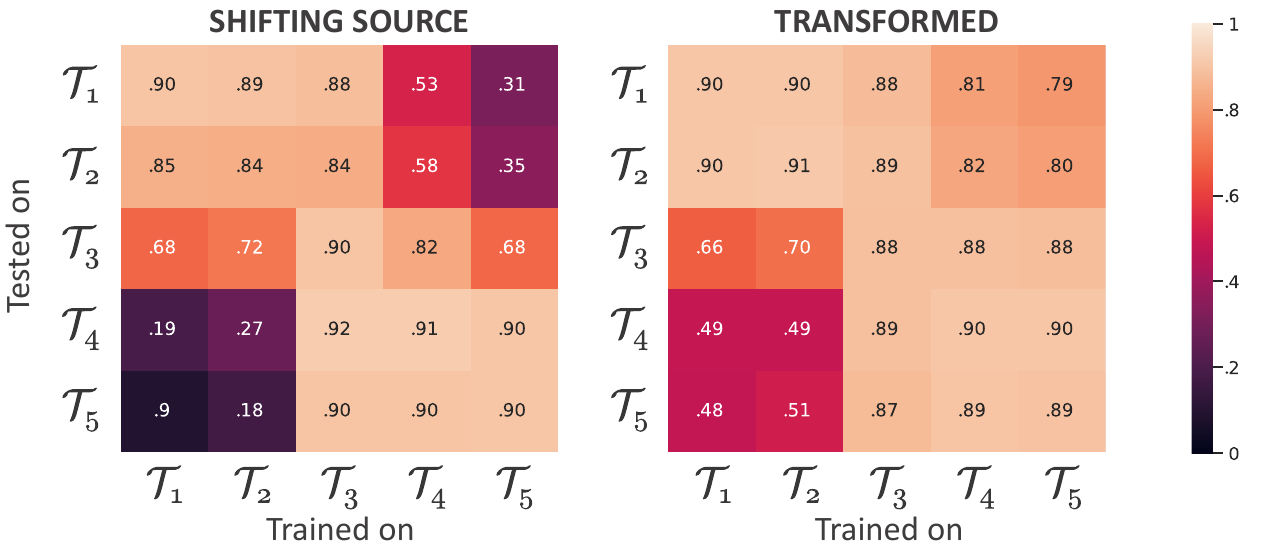}
\caption{Base transferability in terms of Dice score of training separate models statically with each task on test data from each task.} \label{fig:static}
\end{figure}